\documentclass[manuscript,nonacm]{acmart}

\usepackage{times}
\usepackage{soul}
\usepackage{url}
\usepackage{comment}

\usepackage{graphicx}
\usepackage{subcaption}

\AtBeginDocument{%
  }

\copyrightyear{2024}
\acmYear{2024}

\begin{document}

\title{Towards Gradient-based Time-Series Explanations through a SpatioTemporal Attention Network}

\author{Min Hun Lee}
\email{mhlee@smu.edu.sg}
\affiliation{%
  \institution{Singapore Management University}
  \city{Singapore}
  \country{Singapore}
}
\renewcommand{\shortauthors}{Lee.}

\settopmatter{printacmref=false}

\begin{abstract}
In this paper, we explore the feasibility of using a transformer-based, spatiotemporal attention network (STAN) for gradient-based time-series explanations. First, we trained the STAN model for video classifications using the global and local views of data and weakly supervised labels on time-series data (i.e. the type of an activity). We then leveraged a gradient-based XAI technique (e.g. saliency map) to identify salient frames of time-series data. According to the experiments using the datasets of four medically relevant activities, the STAN model demonstrated its potential to identify important frames of videos. 
\end{abstract}

\maketitle

\section{Introduction}
Recent studies showed the competitive performance of artificial intelligence (AI) models on decision-making tasks \cite{rajpurkar2022ai}. However, it is not desirable to apply AI fully autonomously as wrong outcomes of AI models in high-stake domains could have serious impacts on people. Regardless of the performance of an AI model, the end-users desire to understand the evidence on the outcome of an AI model \cite{wang2020should}. A growing body of research investigates how to generate explanations of an AI model and augment user's decision-making tasks \cite{cai2019human,lee2021human,rojat2021explainable}. 

Researchers have explored various techniques to make AI interpretable and explainable \cite{lakkaraju2020explaining}. These explainable AI techniques can be broadly categorized into inherently interpretable models (e.g. rule-based models or linear regressions) whose internal mechanisms are directly interpreted and post-hoc explainable AI (XAI) techniques
that provide explanations of a complex algorithm. 
Post-hoc explanations generate approximations of an AI model outcome by producing understandable information through saliency maps of an AI model outcome for image data \cite{simonyan2013deep}.  Specifically, these saliency maps calculate the gradient of the loss function with respect to the input pixels for the class of an AI model outcome. Thus, these saliency maps are useful to provide insights into which part of an image is considered important for the model outputs.

While saliency maps are traditionally for tabular or image data, there has been limited understanding of how XAI techniques can be applied to time-series data. Some studies have explored adapting saliency map methods for time series data \cite{selvaraju2017grad,goodfellow2018towards,rojat2021explainable}. However, these previous works focus on exploring saliency maps as an explanation tool for a model’s output through qualitative reviews. For instance, after generating sequences of saliency maps of a video, these work overlaid these maps on the input data and checked qualitatively whether the highlighted areas of time-series data are relevant to a model output or not \cite{samek2017explainable}. It remains unclear how well saliency maps can be used to identify salient frames in time-series data.

In this paper, we present a tranformer-based spatiotemporal attention network (STAN) for gradient-based time-series explanations (Figure \ref{fig:architecture}). Our approach first trains a transformer-based model for video classification \cite{li2022uniformer} using the dataset of four medically relevant activities with global and local (i.e. region of interest) views of videos and weakly supervised labels (i.e. the type of an activity in a video). We then utilize a gradient-based explainable AI technique (XAI), saliency map (e.g. smoothgrad) \cite{simonyan2013deep,lee2023exploring} to compute the gradients of the loss function with respect to input data and identify important frames of a video for time-series explanations. After implementing our approach, we utilized the frame-level annotations (i.e. when an activity occurs in video frames) to evaluate the feasibility of our approach for time-series explanations. During the experiments, we studied the effect of using global and local views, three widely used gradient-based methods (i.e. vallina gradient \cite{simonyan2013deep}, smooth grad \cite{smilkov2017smoothgrad}, and gradcam \cite{selvaraju2017grad}), using short and long sequences of video frames. Our results showed that leveraging global and local views improved the performance of both video classification and time-series explanations than using only global or local views. Our STAN model with the global and local views and smoothgrad method achieved an 86.68 F1-score to identify important frames of short videos (i.e. 20 frames), which is a 4.68 F1-score higher than the CNN model with only global views that achieves 82.04 F1-score. However, we also found the limitation of our STAN model on a long sequence of video frames (i.e. 394 frames): the STAN model with the global and local views achieved 87.21 F1-score while the CNN model with the global and local views achieved 95.12 F1-score. We further discussed the potential and limitations of using a transformer-based attention model and gradient-based methods for time-series explanations. 

\section{Related Work}

As it becomes critical to provide explanations of complex AI models to build  trust with the end user, various explainable AI techniques have been studied to provide evidence on the outcome of an AI model \cite{lakkaraju2020explaining}. These XAI techniques can be categorized into inherently interpretable models, such as linear regression models and decision trees, and post-hoc methods that generate approximate of the model's decision by producing understandable representation, such as relevant examples or saliency maps \cite{lakkaraju2020explaining}. 
Prototype/example-based approaches aim to provide a representative sample or region in the latent space \cite{kim2018interpretability}. 
Saliency maps highlight the regions contributing to the outcome of an AI model \cite{lakkaraju2020explaining}. Even if XAI techniques have been widely explored for tabular or image data \cite{shimoda2016distinct}, there have been little explorations on how to explain time-series data \cite{rojat2021explainable,tonekaboni2020went}. 

Along this line, some recent works of XAI have explored the feasibility of using saliency maps for time-series \cite{parvatharaju2021learning,tonekaboni2020went,ismail2020benchmarking}. For example, \cite{ismail2020benchmarking} generated the synthetic dataset to capture temporal-spatial aspects and discussed the limitation of saliency methods on recurrent neural networks, temporal convolutional networks, and transformer models to capture temporal-spatial aspects. Building upon these previous studies, we contributed to the empirical study on exploring the feasibility of using saliency maps with the dataset of four medically relevant activities and the state-of-the-art spatiotemporal attention model for video classification \cite{li2022uniformer} that leverage 3D convolution and global and local spatiotemporal self-attention for effective spatiotemporal representation learning.

\section{Methods}
\begin{figure*}[htp!]
\centering
\includegraphics[width=1.0\textwidth]{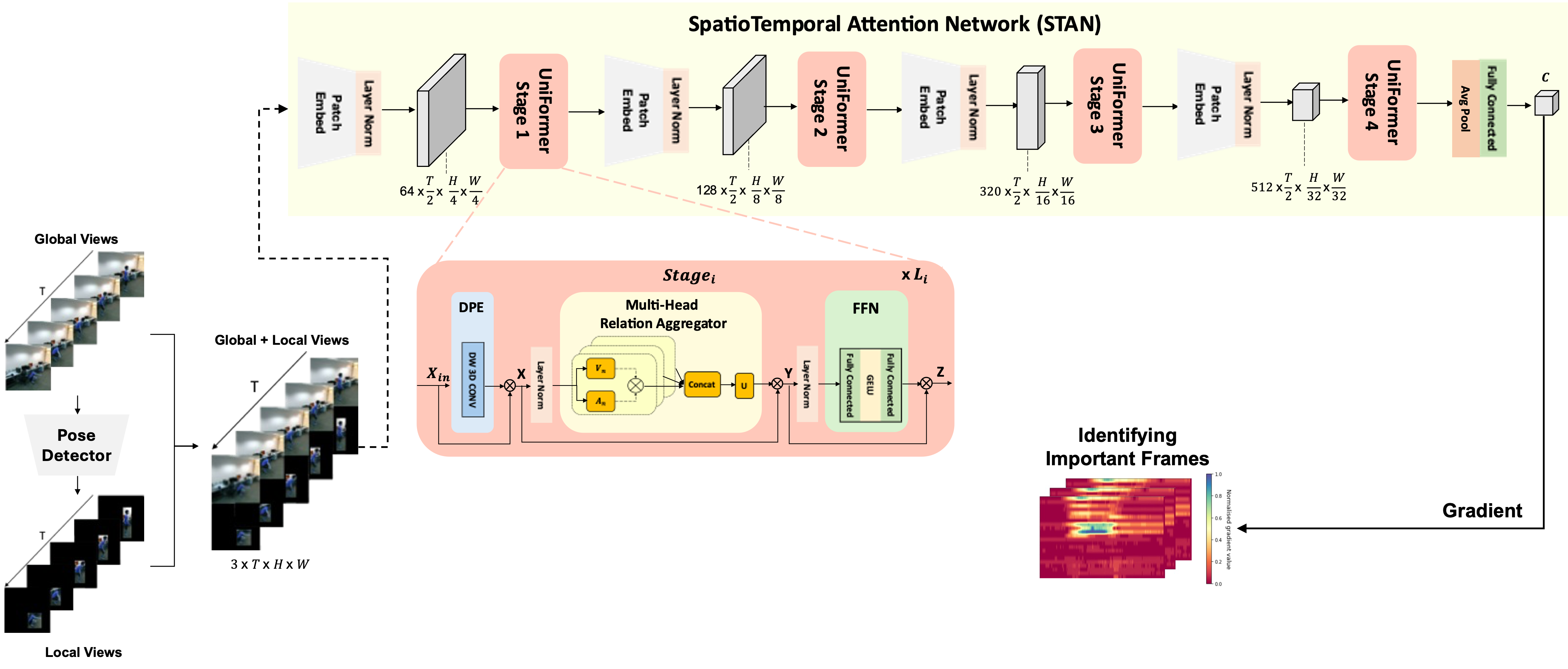}
\caption{Overall flow diagram of a SpatioTemporal Attention Network (STAN) for gradient-based time-series explanation: our approach first learns a transformer-based spatiotemporal attentional network for video classification using global and local views and weakly supervised labels. The STAN consists of four transformer-based attention stages. Given our STAN model, we compute the gradients of a video classification and leverage the gradient scores to identify important frames of a video.}~\label{fig:architecture}
\end{figure*}

In this section, we describe our tranformer-based SpatioTemporal Attention Network (STAN) for gradient-based time-series explanations (Figure \ref{fig:architecture}). Specifically, our approach is composed of 1) learning a global-local spatiotemporal attention-based video classification model with weakly supervised labels and 2) utilizing a gradient-based explainable AI method to identify the important frames of a video. 

\subsection{SpatioTemporal Attention Network (STAN)}\label{sect:glstan}
We first describe the overall architecture of the STAN for video classification and then elaborate on its basic block of Unified transFormer (UniFormer) \cite{li2022uniformer} for efficient and effective spatiotemporal representation learning.

The STAN model is a weakly supervised video classification model that takes a video clip $\mathcal{X} \in \mathbb{R}^{3 \times T\times H\times W}$ as input and classifies the category of a video as an output (e.g. the type of an activity - falling down, headache, chest pain, or compensation motion while rehabilitation - Figure \ref{fig:sample_datasets}). $T$ indicates the number of frames with 3 channels of $H \times W$ pixels in a video. 

The STAN model includes the four stages of UniFormer blocks \cite{li2022uniformer} and applies a hierarchical design \cite{yang2021focal,fan2021multiscale} to generate feature maps whose resolutions are decreasing from early to late stages for processing high-resolution videos. Before the first stage, we applied a $3\times4\times 4$ convolution with the stride of $2\times4\times4$ to downsample and partition a video, resulting in $\frac{T}{2}\times\frac{H}{4}\times\frac{W}{4}$ visual tokens with 64 feature channel dimension. Before other stages, we applied a $1\times2\times 2$ convolution with the stride of $1\times2\times2$, resulting in $\frac{T}{2}\times\frac{H}{8}\times\frac{W}{8}$ visual tokens with 128 feature channel, $\frac{T}{2}\times\frac{H}{16}\times\frac{W}{16}$ visual tokens with 320 feature channel, and $\frac{T}{2}\times\frac{H}{32}\times\frac{W}{32}$ visual tokens with 512 feature channel respectively. At each stage, we utilized a transformer-based UniFormer block with two different attention mechanisms: a local or a global UniFormer block \cite{li2022uniformer}. In the first two stages, we utilized a local UniFormer block to reduce the short-term spatiotemporal redundancy. In the last two stages, we applied a global UniFormer block to capture long-range dependency and context information in a video. These four stages include three, four, eight, and three UniFormer blocks respectively. Finally, the output of the last stage is applied with an average polling followed by a fully connected layer to generate the final classification outputs. 

\subsubsection{Global \& Local Views}
Given a video, our approach leverages global views of image frames ({global}), local views of image frames ({local}), in which we locate the region of interest (ROI) within the frame (e.g. the location of a human who performs an activity) and masked out non-ROI pixels, or global and local views that merge original frames with the ROI frames. For the context of our study on human activity recognition, we utilized the pose estimation model \cite{cao2017realtime} to identify the body joint coordinates of a person and generate local views of image frames. As an AI/ML model might learn undesirable correlations during training data (e.g. identifying snow in the background as important pixels to classify Eskimo dogs, huskies in an image classification task) \cite{ribeiro2016should}, this work hypothesizes that leveraging local views with the ROI might induce an AI/ML model to learn desirable correlations from data and improve its performance. 

\subsubsection{UniFormer Block}
The UniFormer block \cite{li2022uniformer} consists of three key modules: 1) dynamic position embedding (DPE), 2) multi-head relation aggregator (MHRA), and feed-forward network (FFN). 

The DPE aims to encode spatiotemporal position information for token representations. As convolutions can provide absolute position information \cite{islam2020much,chu2021conditional}, the DPE applies 3D depthwise convolution with zero paddings that help the tokens on the borders to be aware of their absolute positions\cite{chu2021conditional,li2022uniformer}.

The MHRA first concatenates resulting normalized tokens weighted by Token Affinity and Value functions and the MHRA can be local or global. The local MHRA aims to learn video representation from the local spatiotemporal context. The local MHRA normalizes the input token with Batch Normalization \cite{ioffe2015batch}. The affinity function is a learnable parameter matrix operated in the local 3D neighborhood to indicate affinity between an anchor token and other tokens 
and the value function implements pointwise convolution (PWConv). The relation aggregator can be considered as applying a depthwise convolution (DWConv) and the final concatenation of all heads can be instantiated as PWConv. Overall, the local MHRA can be considered as applying PWConv, depthwise convolution (DWConv), and PWConv similar to the MobileNet block \cite{sandler2018mobilenetv2}. The global MHRA aims to capture long-term token dependency and follows the design of self-attention in a transformer \cite{dosovitskiy2020image,tolstikhin2021mlp}. The token affinity function of the global MHRA computes content similarity among all the tokens in the global 3D tube while query and key are linear transformations. The input tokens are normalized by the Layer Normalization \cite{ba2016layer}.

The FFN contains two fully connected layers, in which the first layer is followed by the Gaussian error linear units (GELU) \cite{hendrycks2016gaussian}.

\subsection{Gradient-based Time-Series Explanations}\label{sect:frame-score}
For identifying important frames of a video, our approach computes the gradient of the video classification score with respect to input data over all frames of a video. We then normalized the gradient scores of all frames into frame-level scores [0, 1] by averaging the scores over channels and aggregating them. 

\begin{figure*}[htp!]
\centering
\begin{subfigure}[t]{0.35\textwidth}
\centering
  \includegraphics[width=1.0\columnwidth]{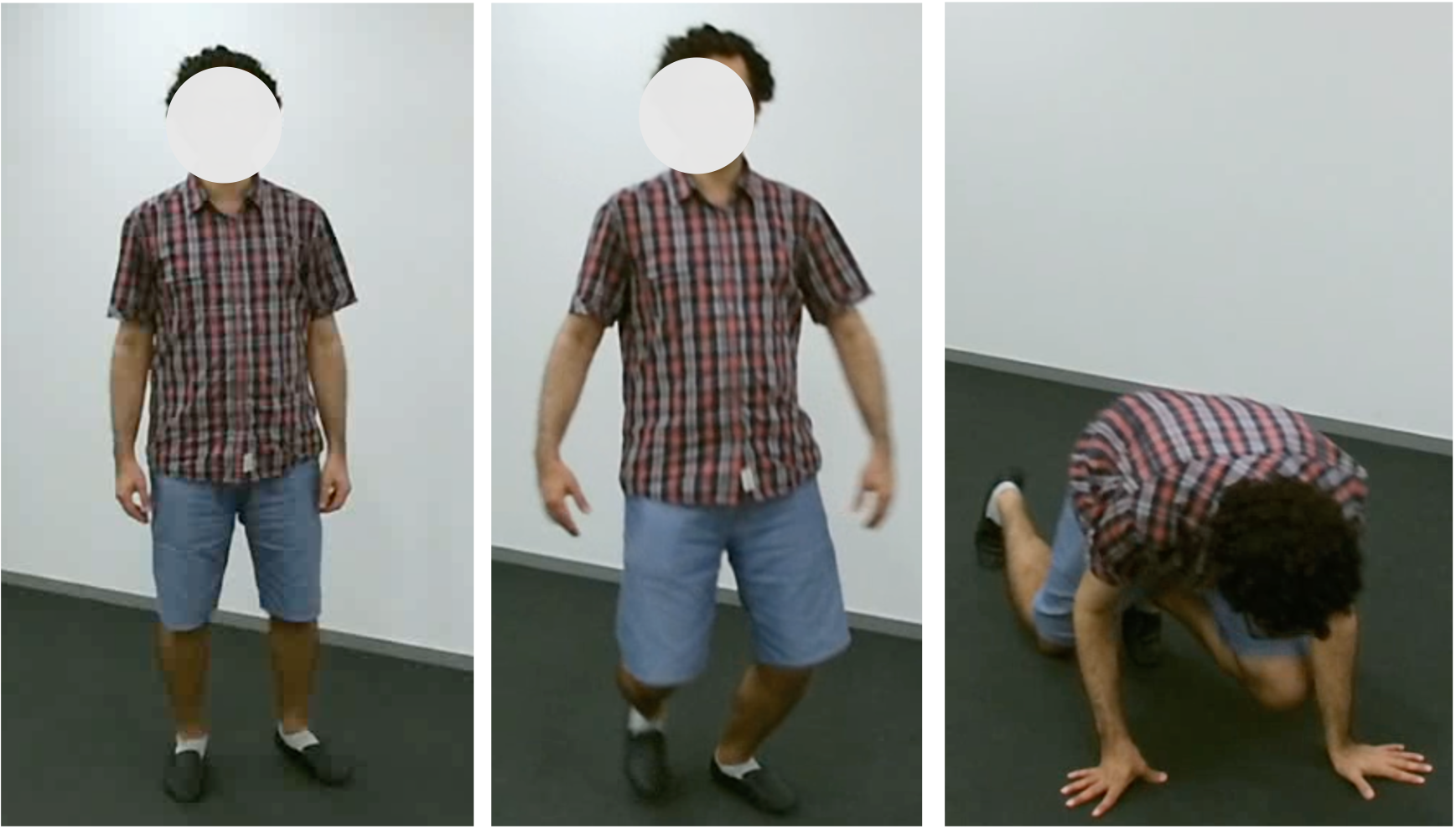}
  \caption{}
  \label{fig:sample_fallingdown}
\end{subfigure}
\begin{subfigure}[t]{0.21\textwidth}
\centering
  \includegraphics[width=1.0\columnwidth]{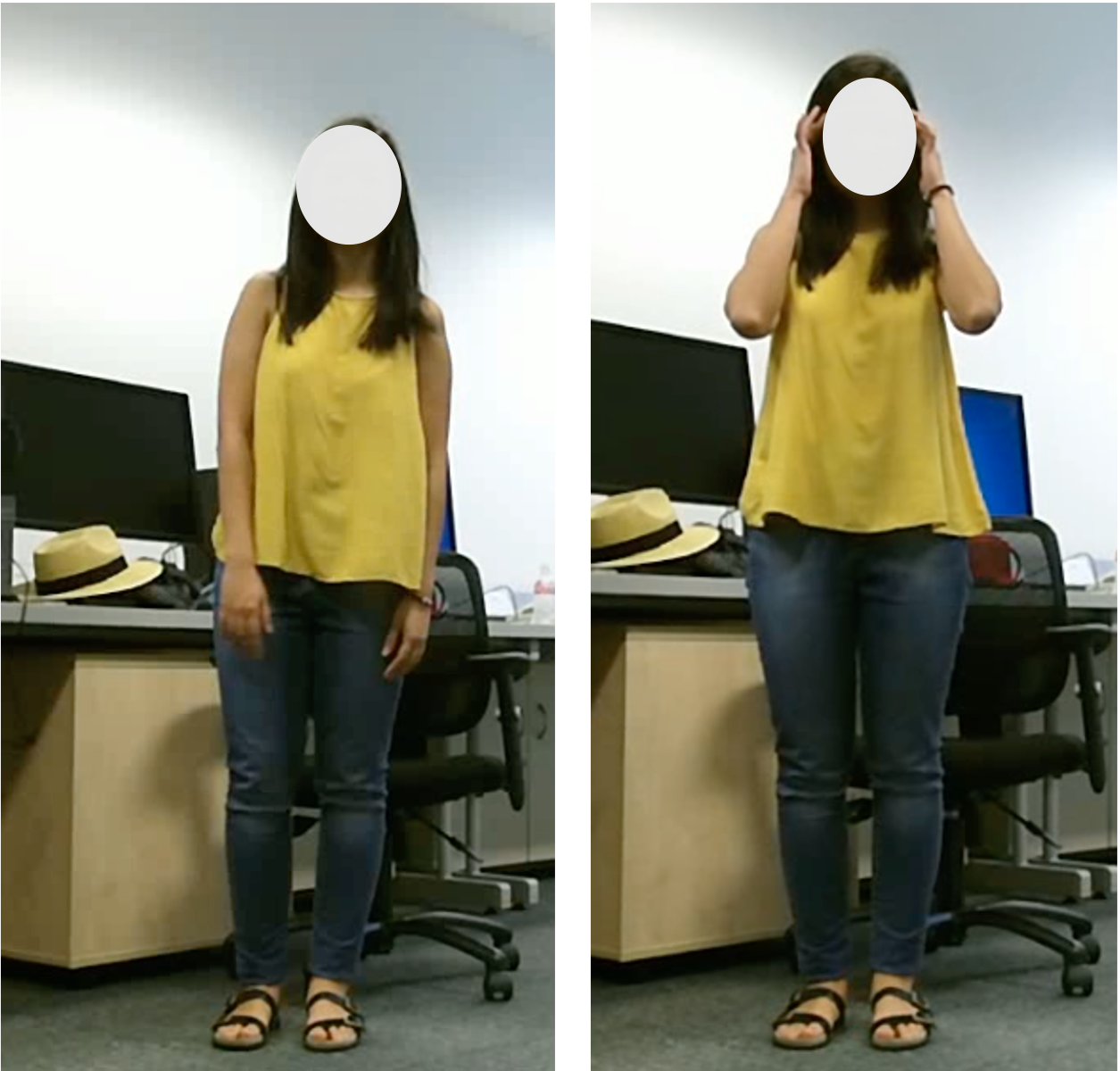}
  \caption{}
  \label{fig:sample_headache}
\end{subfigure}
\begin{subfigure}[t]{0.22\textwidth}
\centering
  \includegraphics[width=1.0\columnwidth]{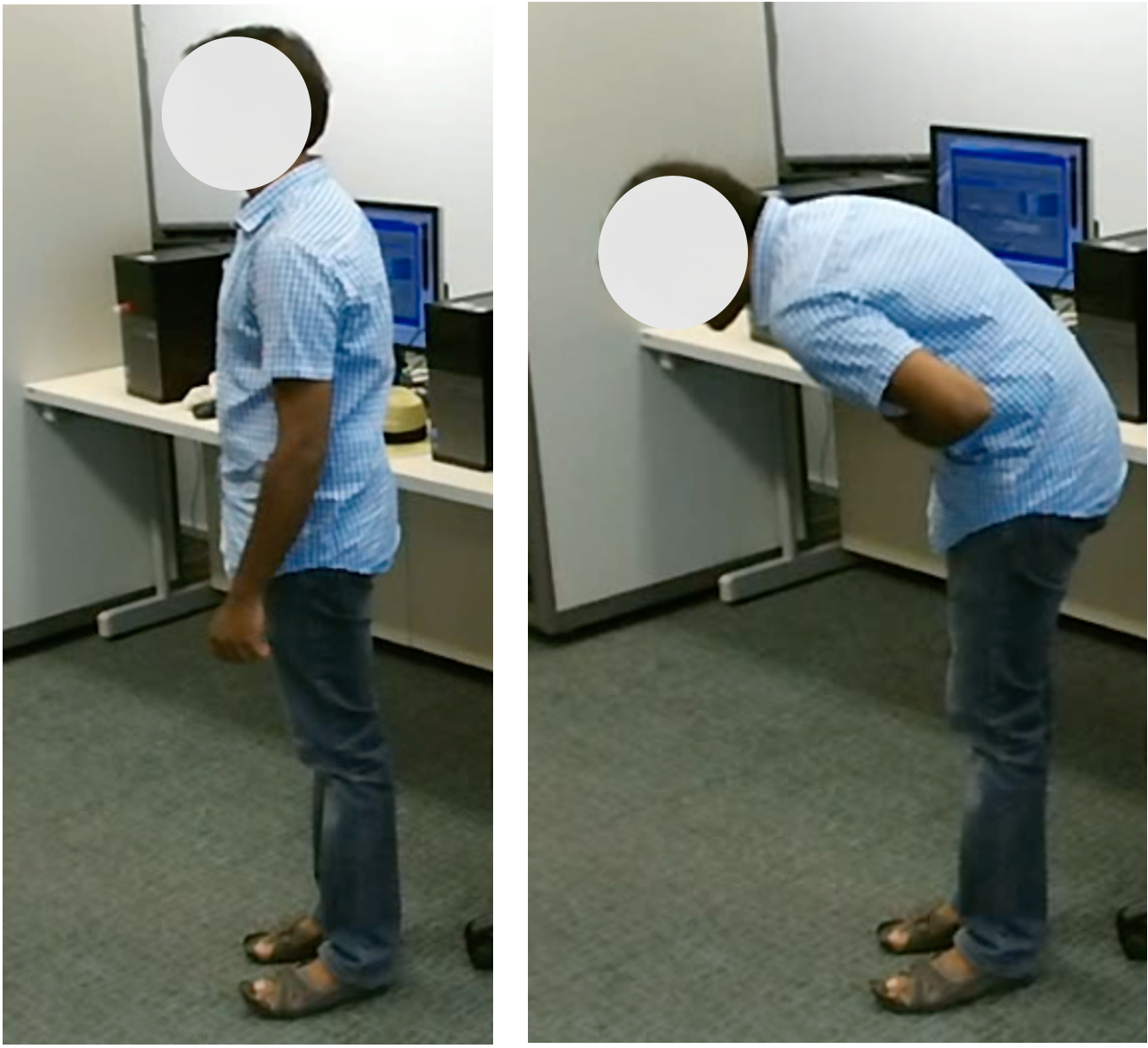}
  \caption{}
  \label{fig:sample_chestpain}
\end{subfigure}
\begin{subfigure}[t]{0.18\textwidth}
\centering
  \includegraphics[width=1.0\columnwidth]{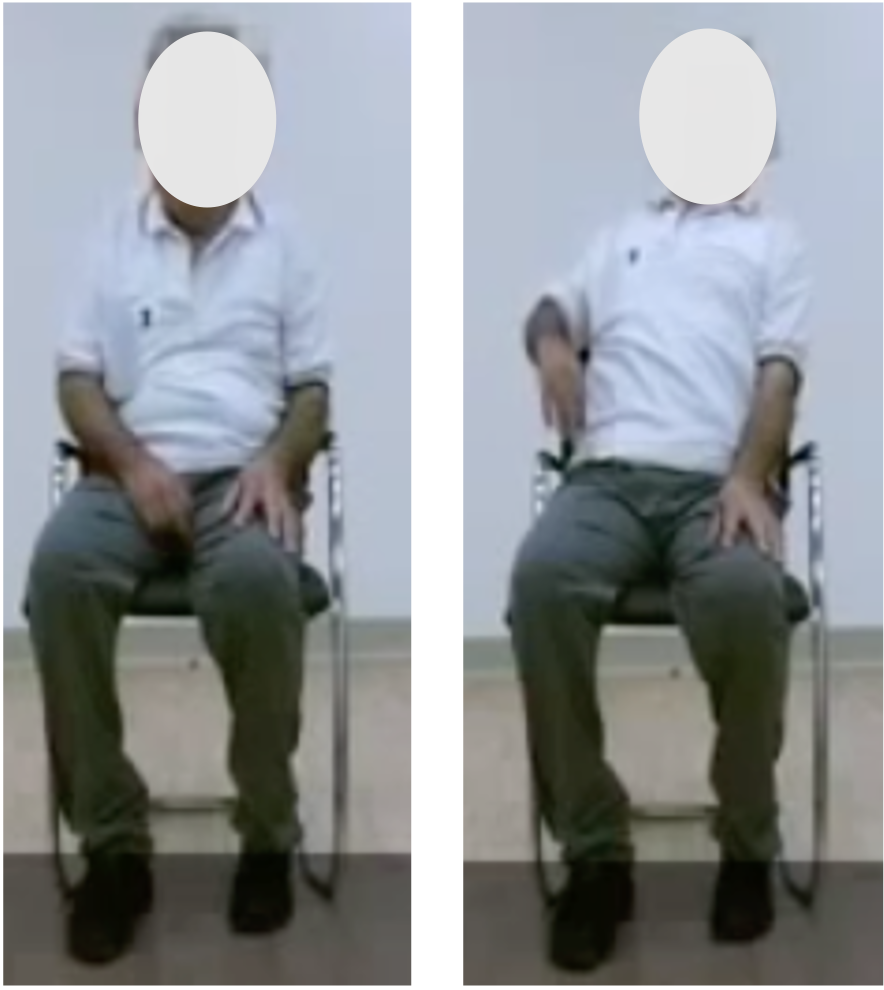}
  \caption{}
  \label{fig:sample_rehab}
\end{subfigure}
\caption{Sample frames of the datasets: (a) Falling Down - Frontal View (b) HeadAche - Frontal View (c) ChestPain - SideView, and (d) Rehabilitation Compensation - Frontal View.}~\label{fig:sample_datasets}
\end{figure*}

We hypothesize that there is a threshold value between the score range [0,1] where scores less than or equal to the threshold score indicate ``not important'' frames, and scores more than the threshold score indicate ``important frames'' for video classification (e.g. where a person perform a specific gesture related to an activity - Figure \ref{fig:sample_datasets}).

We explored three gradient-based methods: 1)  Vanilla Gradient \cite{simonyan2013deep} that computes the gradient of the loss function for the class with respect to input data, 2) SmoothGrad \cite{smilkov2017smoothgrad} that generates multiple inputs by adding a noise to input and computes an average of the pixel attribution maps to make gradient scores less noisy, and 3) Gradient-weighted Class Activation Map \cite{selvaraju2017grad} that back-propagates the gradient to the last convolutional layer to produce a saliency map for highlighting important regions of an image.

\section{Experiments}
We conducted experiments to evaluate the effectiveness of our STAN for 1) video classification and 2) gradient-based time-series explanations to identify important frames of a video. In this work, we focus on the context of video analysis on four human activities that are relevant to healthcare. 

\subsection{Datasets}
For the context of our study, we utilized the NTU RGB+D dataset \cite{shahroudy2016ntu} and the stroke rehabilitation dataset \cite{lee2020towards}. 

The NTU RGB+D dataset \cite{shahroudy2016ntu} contains 114,480 video samples of 60 activity classes. Among 60 activity classes, we focused on utilizing a subset of three medically relevant activity classes: falling down (Figure \ref{fig:sample_fallingdown}), headache (\ref{fig:sample_headache}), and chest pain (Figure \ref{fig:sample_chestpain}). Overall, each medically relevant activity includes 948 videos respectively. To evaluate our proposed approach of gradient-based time-series explanations, we selected 900 samples of 3 medically relevant activities and generated frame-level annotations which frames a medically relevant activity occurs or not. 

The stroke rehabilitation dataset \cite{lee2020towards} includes 300 videos of 15 post-stroke survivors performing a rehabilitation exercise. Some post-stroke survivors with limited functional abilities involved compensatory motions (Figure \ref{fig:sample_rehab}), in which they might use unnecessary joint movements, such as leaning the trunk backward. The dataset includes two types of labels: (1) an overall label of whether an exercise involve a compensatory motion and (2) frame-level labels on when a compensatory motion occurs or not. 

\subsection{Video Classification Models}
For training our STAN model, we leveraged the unified transformer model \cite{li2022uniformer} that was pre-trained with Kinetics-400 \cite{carreira2017quo} and then further trained the model using the NTU RGB+D dataset and the stroke rehabilitation dataset with weakly supervised labels (i.e. types of an activity). We applied 10-fold cross-validation for training our model with 2,844 videos of the medically relevant activities and applied leave-one-patient-out cross-validation for training our model with 300 videos of the stroke rehabilitation exercises. We utilized AdamW optimizer with cosine learning
rate schedule \cite{loshchilov2016sgdr,loshchilov2018fixing} and the batch size of 64 to train the STAN model. 

As the purpose of this study is to demonstrate the feasibility of using the gradients of a transformer-based video classification model for time-series explanation and the previous study \cite{li2022uniformer} showed the competitive performance of a unified transformer model for activity video classifications, we implemented basic baseline models video classification models: 1) ResNet50 model with a fully connected layer (FCL), 2) ResNet50 model with a convolutional layer followed by a maxpooling layer and two fully connected layers (CNN), three LSTM layers with two fully connected layers (RNN $|$ LSTM). For training these baseline models, we utilized the final linear layer of the ResNet50 model that was/was not pre-trained with the Kinetics 400 dataset to extract a feature vector of an image frame. For FCL and CNN, we concatenated these feature vectors of image frames into patched images as an input for the FCL and CNN models and utilized a sequence of feature vectors from image frames as an input for RNN $|$ LSTM model. 
In addition, we studied the effect of using global, local, and global + local views of image frames for input data for video classification models.

\subsection{Gradient-Based Time-Series Explanations}
Another experiment is to study how well our gradient-based time-series explanations can identify important frames of a video (e.g. when a specific activity occurs). Once we trained our STAN model, we utilized training data to empirically select a threshold value and evaluate the selected threshold value on test data to identify important frames of a video. 

For the experiment, we utilized the dataset of four medically relevant activities along with frame-level annotations and compared the performance to identify important frames of a video over several aspects using an F1-score. These evaluation aspects include (i) types of a video classification model (i.e. a CNN-based video classification model with input data of patched images vs a transformer-based, attention network), (ii) types of data views (i.e. global, local, global + local), (iii) methods of computing gradients, and (iv) a short (i.e. 20 frames) and a long sequence (i.e. 394 frames) of image frames.

\section{Results}
\subsection{Video Classification Models}
Table \ref{tab:results_videoclassifers} summarizes the performance of various approaches for video classifications.
Overall, our STAN that leverages the unified transformer (UniFormer) blocks pre-trained with Kinetics and global + local views achieved the highest performance of video classification with 96.49 F1-score to classify the type of an activity.

\begin{table}[htp]
\centering
\resizebox{0.75\columnwidth}{!}{%
\begin{tabular}{llcll} \toprule
\multicolumn{3}{c}{\textbf{Model}}                       & \multicolumn{1}{c}{\textbf{Data}} & \multicolumn{1}{c}{\textbf{Overall}} \\ \midrule
FCL        & ResNet50  & -        & Global         & 80.87 \\
FCL        & ResNet50  & Kinetics & Global         & 81.48 \\
CNN        & ResNet50  & Kinetics & Global         & 82.43 \\
RNN $|$ LSTM & ResNet50  & Kinetics & Global         & 83.61 \\
STAN      & Uniformer & Kinetics & Global         & 95.43 \\ \midrule
FCL        & ResNet50  & Kinetics & Local          & 74.20 \\
CNN        & ResNet50  & Kinetics & Local          & 74.20 \\
STAN      & Uniformer & Kinetics & Local          & 95.61 \\ \midrule
FCL        & ResNet50  & Kinetics & Global + Local & 90.63 \\
CNN        & ResNet50  & Kinetics & Global + Local & 90.39 \\
\textbf{STAN} & \textbf{UniFormer} & \textbf{Kinetics} & \textbf{Global + Local}           & \textbf{96.49} \\ \bottomrule                     
\end{tabular}%
}
\caption{Performance of Weakly Supervised Video Classification Models with various settings (Backbones: ResNet50 or Uniformer, with/without pretraining with the Kinematics Dataset, with the dataset of Global, Local, and Global + Local views). Our STAN with Uniformer backbones and Global + Local data achieved the highest performance.}
\label{tab:results_videoclassifers}
\end{table}

When we compared the performance of the FCL model with the ResNet50 (80.87 F1-score) and that of the FCL model with the pre-trained ResNet50 with the Kinetics 400 dataset (81.48), the latter model achieved a 1.48 higher F1-score, which demonstrated the positive effect of pre-training with Kinetics. Among various models that leverage global views of videos, we found that the transformer-based attention model (STAN) with global views achieved a 95.43 F1-score, which outperformed other models with the data of global views. Similarly, the STAN with local views achieved a 95.61 F1-score, which outperformed other models with the local views. Compared to the STAN with global views, the STAN with local views improved its performance slightly by 0.18 F1-score. When both global and local views are utilized, the STAN model improved its performance to 96.49 F1-score. The STAN with global and local views outperformed other models with global+local views, which showed the benefits of applying local and global multi-head attention for effective spatiotemporal learning.

\subsection{Gradient-Based Time-Series Explanations}
In this section, we summarize the performance of gradient-based time-series explanations for identifying important frames of videos through various approaches. In addition, we describe the performance of gradient time-series explanations over a short frame (i.e. 20 frames) and a long frame (i.e. 394 frames). 

Table \ref{tab:results_frames_vanilagrad}, Table \ref{tab:results_frames_smooth}, and Table \ref{tab:results_frames_GradCam} include the results of various methods using the vanilla gradient, smoothgrad, and gradcam respectively. Similar to the video classification, we observed that the STAN with global + local views using vanilla grad (i.e. 86.43 F1-score) and smoothgrad (i.e. 86.68 F1-score) achieved the highest performance of identifying important frames of videos compared to the CNN-based model with different views of data. The application of smoothgrad led to a slight performance improvement from 86.43 to 86.68 F1-score than using vanilla grad. However, when it comes to using a gradcam method, we observed that the CNN-based model with global + local views achieved the best performance of 82.22 F1-score, which is higher than that of the STAN with global + local views with 80.60 F1-score.

\begin{table*}[htp]
\centering
\resizebox{0.8\linewidth}{!}{%
\begin{tabular}{lccccc} \toprule
\multicolumn{1}{c}{\textbf{VanillaGrad}} & \textbf{FallingDown} & \textbf{HeadAche} & \textbf{ChestPain} & \textbf{\begin{tabular}[c]{@{}c@{}}Rehabilitation\\ Compensation\end{tabular}} & \textbf{Overall} \\ \midrule
CNN - Global         & 81.64 & 85.94          & 81.67 & 79.11          & 82.09 \\
CNN - Local          & 80.64 & 87.26          & 82.06 & 79.35          & 82.33 \\
CNN - Global + Local & 83.07 & 86.97 & 81.77 & 79.76 & 82.89 \\ \midrule
STAN - Global      & 85.22 & 91.06          & 83.99 & 83.20          & 85.87 \\
STAN - Local       & 85.25 & 89.45          & 85.27 & 83.32          & 85.82 \\
STAN - Global + Local                  & \textbf{85.81}       & \textbf{91.25}    & \textbf{85.30}     & \textbf{83.35} & \textbf{86.43}  \\ \bottomrule
\end{tabular}%
}
\caption{Results of identifying important frames using VanillaGrad on Falling, HeadAche, ChestPain, and Rehab Compensation datasets: Our STAN - Global + Local achieved the highest performance compared to other approaches}
\label{tab:results_frames_vanilagrad}
\end{table*}

\begin{table*}[htp]
\centering
\resizebox{0.8\linewidth}{!}{%
\begin{tabular}{lccccc} \toprule
\multicolumn{1}{c}{\textbf{SmoothGrad}} &
  \multicolumn{1}{c}{\textbf{FallingDown}} &
  \multicolumn{1}{c}{\textbf{HeadAche}} &
  \multicolumn{1}{c}{\textbf{ChestPain}} &
  \multicolumn{1}{c}{\textbf{\textbf{\begin{tabular}[c]{@{}c@{}}Rehabilitation\\ Compensation\end{tabular}} }} &
  \multicolumn{1}{c}{\textbf{Overall}} \\ \midrule
CNN - Global            & 80.37          & 86.53          & 81.92          & 79.32          & 82.04          \\
CNN - Local             & 81.14          & 86.91          & 81.81          & 79.44          & 82.33          \\
CNN - Global + Local    & 83.02          & 88.12          & 82.39          & 79.89          & 83.36          \\ \midrule
STAN - Global         & 85.30          & 91.08          & 85.11          & 83.25          & 86.19          \\
STAN - Local          & 85.50          & 89.64          & 85.34          & 83.55          & 86.01          \\
STAN - Global + Local & \textbf{85.83} & \textbf{91.33} & \textbf{85.88} & \textbf{83.69} & \textbf{86.68} \\ \bottomrule
\end{tabular}%
}
\caption{Results of identifying important frames using SmoothGrad on Falling, HeadAche, ChestPain, and Rehab Compensation datasets: Our STAN - Global + Local achieved the highest performance compared to other approaches}
\label{tab:results_frames_smooth}
\end{table*}

\begin{table*}[htp]
\centering
\resizebox{0.8\linewidth}{!}{%
\begin{tabular}{lccccc}\toprule 
\multicolumn{1}{c}{\textbf{GradCam}} &
  \multicolumn{1}{c}{\textbf{FallingDown}} &
  \multicolumn{1}{c}{\textbf{HeadAche}} &
  \multicolumn{1}{c}{\textbf{ChestPain}} &
  \multicolumn{1}{c}{\textbf{\textbf{\begin{tabular}[c]{@{}c@{}}Rehabilitation\\ Compensation\end{tabular}} }} &
  \multicolumn{1}{c}{\textbf{Overall}} \\ \midrule
CNN - Global            & 82.39          & 85.46          & 80.29          & 77.05          & 81.30          \\
CNN - Local             & 82.00          & \textbf{87.20} & \textbf{81.60} & 77.50          & 82.08          \\
CNN - Global + Local    & \textbf{83.04} & 86.92          & 81.37          & 77.53          & \textbf{82.22} \\ \midrule
STAN - Global         & 78.74          & 85.03          & 79.44          & 78.66          & 80.47          \\
STAN - Local          & 78.89          & 78.91          & 79.46          & 78.91          & 79.04          \\
STAN - Global + Local & 78.91          & 85.05          & 79.46          & \textbf{78.98} & 80.60     \\ \bottomrule    
\end{tabular}%
}
\caption{Results of identifying important frames using GradCam Falling, HeadAche, ChestPain, and Rehab Compensation datasets: Our STAN - Global + Local achieved the highest performance compared to other approaches only on the rehabilitation compensation dataset while CNN-based model outperformed our STAN - Global + Local on FallingDown, HeadAche, ChestPain}
\label{tab:results_frames_GradCam}
\end{table*}

\begin{table*}[htp]
\centering
\resizebox{0.8\linewidth}{!}{%
\begin{tabular}{lcccccc} \toprule
\multicolumn{1}{c}{}    & \multicolumn{2}{c}{\textbf{Vanilla}} & \multicolumn{2}{c}{\textbf{SmoothGrad}} & \multicolumn{2}{c}{\textbf{GradCam}} \\ \midrule
\multicolumn{1}{c}{\textbf{}} &
  \textbf{\begin{tabular}[c]{@{}c@{}}Short Seq\\ 20 frames\end{tabular}} &
  \textbf{\begin{tabular}[c]{@{}c@{}}Long Seq\\ All\end{tabular}} &
  \textbf{\begin{tabular}[c]{@{}c@{}}Short Seq\\ 20 frames\end{tabular}} &
  \textbf{\begin{tabular}[c]{@{}c@{}}Long Seq\\ All\end{tabular}} &
  \textbf{\begin{tabular}[c]{@{}c@{}}Short Seq\\ 20 frames\end{tabular}} &
  \textbf{\begin{tabular}[c]{@{}c@{}}Long Seq\\ All\end{tabular}} \\ \midrule
CNN - Global            & 79.11             & 94.17            & 79.32              & 94.92              & 77.05             & 94.81            \\
CNN - Local             & 79.35             & 94.22            & 79.44              & 95.39              & 77.50             & 95.01            \\
CNN - Global + Local    & 79.76             & \textbf{94.56}   & 79.89              & \textbf{95.60}     & 77.53             & \textbf{95.12}   \\ \midrule
STAN - Global         & 83.20             & 87.30            & 83.25              & 87.69              & 78.66             & 87.21            \\
STAN - Local          & 83.32             & 87.46            & 83.55              & 87.50              & 78.91             & 87.19            \\
STAN - Global + Local & \textbf{83.35}    & {87.67}   & \textbf{83.69}     & {87.80}     & \textbf{78.98}    & 87.21  \\ \bottomrule       
\end{tabular}%
}
\caption{Results of identifying important frames using short and long sequences using Rehab Compensation dataset: Our STAN - Global + Local achieved the highest performance on the dataset with a short sequence (20 frames) while CNN-based Global + Local model outperformed our STAN - Global + Local on the dataset with a long sequence (all).}
\label{tab:results_frames_shortlong}
\end{table*}

\begin{figure*}[htp!]
\centering
\begin{subfigure}[t]{1.0\textwidth}
\centering
  \includegraphics[width=1.0\columnwidth]{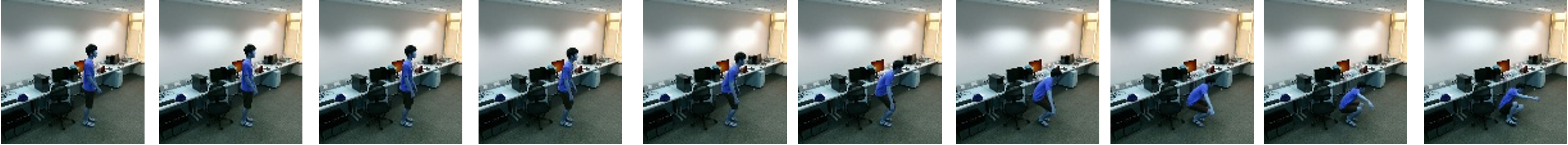}
  \caption{}
  \label{fig:visualization_input}
\end{subfigure}
\begin{subfigure}[t]{1.0\textwidth}
\centering
  \includegraphics[width=1.0\columnwidth]{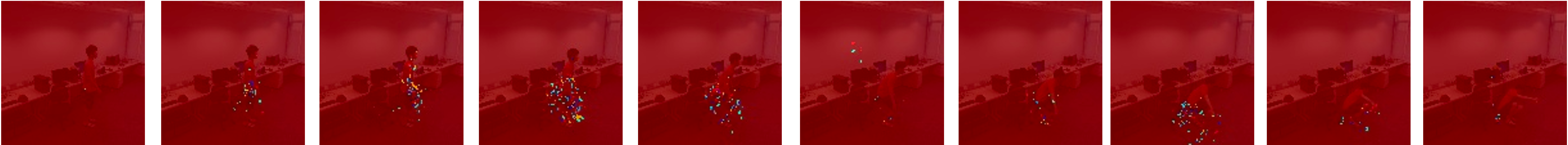}
  \caption{}
  \label{fig:visualization_gradient}
\end{subfigure}
\begin{subfigure}[t]{1.0\textwidth}
\centering
  \includegraphics[width=1.0\columnwidth]{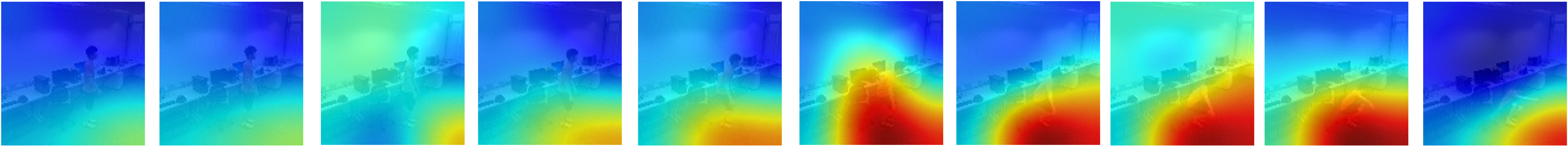}
  \caption{}
  \label{fig:visualization_gradcam}
\end{subfigure}
\caption{(a) Original Inputs (b) VanillaGradient (c) GradCam using the STAN model with global+local views of data. Overall, the model with the vanilla gradient highlighted the focused body areas that are relevant to falling down while the model with gradcam tended to have more diffused attention and some attention on background areas}~\label{fig:sample-exercise-saliencymap}
\end{figure*}

Table \ref{tab:results_frames_shortlong} summarizes the performance of various approaches using video classification models with different gradient methods and data views over a short sequence of data (i.e. 20 frames) and a long sequence of data (i.e. 394 frames). Overall, the results showed that the models with the global + local views achieved higher or equivalent performance than the model with only global or local views. 

When it comes to a short sequence of data, the STAN with global + local views achieved the highest performance of 83.35 F1-score using vanilla gradient, 83.69 F1-score using smoothgrad, and 78.98 F1-score using gradcam. 
For a long sequence of data, the CNN-based model with global + local views achieved the highest performance of 94.56 F1-score using vanilla gradient, 95.60 F1-score using smoothgrad, and 95.12 F1-score using gradcam respectively.

\section{Discussion}
In this work, we showed the feasibility of a transformer-based, spatiotemporal attention network (STAN) with global and local views and a gradient-based explainable AI technique for time-series explanations (i.e. identifying important frames of a video). Our results of video classification models (Table \ref{tab:results_videoclassifers}) and time-series explanations through various gradient methods (Table \ref{tab:results_frames_vanilagrad}, \ref{tab:results_frames_smooth}, \ref{tab:results_frames_GradCam}, and \ref{tab:results_frames_shortlong}) showed \textbf{the benefit of using global and local views for both video classification and time-series explanations} as the models with global + local views performed better or equivalent to other models using only global or local view.

As our STAN model with global+local views achieved the highest performance on video classification tasks, our transformer-based STAN model with gradient methods also achieved the competitive performance of time-series explanations contrary to the finding of \cite{ismail2020benchmarking} that describes the limitation of saliency maps to capture temporal-spatial aspects. Thus, our study demonstrated the potential of a transformer-based global and local attention model for time-series explanations with some limitations.

Even if the STAN model with global+local views using vanilla grad or smoothgrad achieved the highest performance of time-series explanations, the STAN model with global+local views using gradcam performed worse than the CNN model with global + local views. When we analyzed the outputs of our STAN model with vanilla gradient and gradcam (Figure \ref{fig:sample-exercise-saliencymap}), we found that our STAN model with vanilla gradient generated more focused attention on relevant body areas over frames than the model with gradcam that contains more diffused and irrelevant attentions over frames. We speculate that as the grad-cam is limited to the inaccurate and diffused heatmap, it may lose signals during continual upsampling and downsampling processes \cite{chattopadhay2018grad} for time-series explanations.

In addition, although the transformer-based, STAN model achieved higher performance of time-series explanations on a short sequence of frames (i.e. 20 frames), the STAN model performed worse on a long sequence of frames (i.e. 394 frames) than the CNN-based model. Due to the computational resources, when we trained the transformer-based STAN model, we sampled only 20 frames of a video. We then augmented the outputs of the sampled frames by the STAN model to the neighborhood samples for time-series explanations over a long sequence of image frames. The STAN model with such augmentations was not as accurate as the CNN-based model that directly computes the frame-level gradient scores more efficiently. Thus, with limited computational resources, we \textbf{recommend using the transformer-based video classification model, STAN on a short sequence of video frames and the CNN-based model with patched image sequences on a long sequence of video frames for time-series explanations}.

As our STAN model with global+local views can achieve decent performance to identify important frames, our approach has the potential to automatically pinpoint an important segment of a video that requires a time-consuming process. However, as this work only explores one transformer-based, attention model for video classification on exercise-relevant datasets, it is necessary to conduct additional studies to demonstrate the generalizability of our approach with other video classification models, gradient-based explainable AI techniques, and data modalities. In addition, our approach with local views has the limitation of requiring a pre-trained object detection model to localize views of a video for extending it to another application beyond activity-related videos.

\section{Conclusion}
In this work, we contributed to an empirical study that explores the feasibility of using a transformer-based, spatiotemporal attention model (STAN) for gradient-based time-series explanations to identify important frames of a video. Specifically, we utilized the datasets of four medical-relevant activities to study the effect of using global and local (region of interest) views, three widely used gradient-based explainable AI (XAI) techniques (i.e. vanilla gradient, smoothgrad, and gradcam), short and long sequences of video frames. Our results showed the benefit of using both global and local views for both video classification and gradient-based time-series explanations. Our STAN model with vanilla gradient or smoothgrad methods achieves around 86 F1-score and demonstrates its potential of time-series explanations to identify important frames of a short video (i.e. 20 frames) while it has limitations on a long video (i.e. 394 frames).

\bibliographystyle{ACM-Reference-Format}
\bibliography{main}


\end{document}